\documentclass[conference]{IEEEtran}
\usepackage{amssymb,amsmath,amsthm,amsfonts,mathrsfs}
\usepackage[hmargin=3cm,vmargin=3.5cm]{geometry}
\usepackage{cite}
\usepackage{hyperref}
\usepackage{color}
\usepackage{algorithmic}
\usepackage{graphicx}
\usepackage{amscd}  
\usepackage[shellescape]{gmp}
\usepackage{textcomp} 
\usepackage{stmaryrd}  
\usepackage{mathtools}
\usepackage{comment} 
\usepackage{multirow} 
\usepackage[all]{xy}
\usepackage[dvipsnames]{xcolor}
\usepackage{soul} 

\def\BibTeX{{\rm B\kern-.05em{\sc i\kern-.025em b}\kern-.08em
    T\kern-.1667em\lower.7ex\hbox{E}\kern-.125emX}}
  
\usepackage{tikz}
\usepackage{tikz-cd}
\usetikzlibrary{decorations.pathmorphing}

\tikzset{snake it/.style={decorate, decoration=snake}}

\hypersetup{
    colorlinks=true,
    citecolor=Brown,
    filecolor=Brown,
    linkcolor=RedViolet,
    urlcolor=WildStrawberry
}

\usepackage{chngcntr}
\counterwithin{figure}{section}

\allowdisplaybreaks

\title{Computational complexity reduction of deep neural networks}

\author{\IEEEauthorblockN{Mee Seong Im}
\IEEEauthorblockA{\textit{Department of Mathematics} \\
\textit{United States Naval Academy}\\
Annapolis, MD 21402  \\
meeseongim@gmail.com}
\and
\IEEEauthorblockN{Venkat Dasari}
\IEEEauthorblockA{\textit{U.S. Army Combat Capabilities Development Command} \\
\textit{U.S. Army Research Laboratory}\\
Aberdeen Proving Ground, MD 21005 \\
venkateswara.r.dasari.civ@mail.mil}
}

\date{\today}

\begin{document}


\def\R{\mathbb R}
\def\Q{\mathbb Q}
\def\Z{\mathbb Z}
\def\N{\mathbb N} 
\def\C{\mathbb C}
\def\S{\mathbb S}
\def\V{\mathbf V}
\def\FF{\mathbb F}
\renewcommand\SS{\ensuremath{\mathbb{S}}}
\def\l{\lbrace}
\def\r{\rbrace}
\def\o{\otimes}
\def\lra{\longrightarrow}
\def\Ext{\mathsf{Ext}}
\def\FF{\mathsf{FF}}
\def\Fl{\mathsf{Fl}}
\def\GL{\mathsf{GL}}
\def\Gr{\mathsf{Gr}}
\def\Hom{\mathsf{Hom}}
\def\Ind{\mathsf{Ind}}
\def\MF{\mathsf{MF}}
\def\HMF{\mathsf{HMF}}
\def\Res{\mathsf{Res}}
\def\SL{\mathsf{SL}}
\def\Syl{\mathsf{Syl}}
\def\Sym{\mathsf{Sym}}
\def\triv{\mathsf{triv}}
\def\Id{\mathsf{Id}}
\def\mc{\mathcal}
\newcommand{\mcR}{{\mathcal R}}
\def\mf{\mathfrak} 
\def\mcC{\mathcal{C}}
\def\mcO{\mathcal{O}}
\def\U{\mathsf{U}}

\def\biases{\mathsf{biases}}
\def\Cat{\mathsf{Cat}}
\def\End{{\mathsf{End}}} 
\def\tr{{\mathrm{tr}}} 
\def\trG{\mathrm{tr}_{\mathsf{Gr}}}  
\def\det{\mathrm{det}}
\def\undbeta{\underline{\beta}}
\def\mcF{\mathcal{F}}
\def\Gal{\mathsf{Gal}}
\def\mchar{\mathrm{char}} 
\def\Fr{\mathrm{Fr}}
\def\mthH{\mathrm{H}} 
\def\Sym{\mathrm{Sym}}
\def\wt{\mathsf{wt}}

\def\one{\mathbf{1}}   
\def\nchar{\mathrm{char}\,}  

\newcommand{\myrightleftarrows}[1]{\mathrel{\substack{\xrightarrow{#1} \\[-.9ex] \xleftarrow{#1}}}}

\def\lra{\longrightarrow}
\def\CC{\mathbb{C}}
\def\kk{\mathbf{k}}  
\def\ukk{\underline{\kk}}  
\def\undx{\underline{x}} 
\def\undy{\underline{y}} 
\def\undz{\underline{z}} 
\def\unda{\underline{a}}
\def\okk{\overline{\kk}}  
\def\okksep{\overline{\kk}^{sep}} 

\def\indf{\mathsf{Ind}} 
\def\resf{\mathsf{Res}} 
\def\gdim{\mathrm{gdim}}  
\def\rk{\mathrm{rk}}
\def\mmat{\mathrm{Mat}}

\def\Pa{\mathsf{Pa}}   
\def\Cob{\mathsf{Cob}} 
\def\Cobtwo{\Cob_2}   
\def\Kob{\mathsf{Kob}}
\def\Cobal{\Cob_{\alpha}}  

\def\cobal{\mathsf{Cob}_{\alpha}} 

\def\udcob{\underline{\mathsf{DCob}}}

\def\dmod{\mathsf{-mod}}   
\def\pmod{\mathsf{-pmod}}    

\newcommand{\oplusop}[1]{{\mathop{\oplus}\limits_{#1}}}
\newcommand{\ang}[1]{\langle #1 \rangle } 
\newcommand{\bbn}[1]{\mathbb{B}^{#1}}

\newcommand{\pseries}[1]{\kk\llbracket #1 \rrbracket}
\newcommand{\rseries}[1]{R\llbracket #1 \rrbracket}
\newcommand{\ovfi}[1]{\overline{F_{#1}}}
\newcommand{\delcatv}[1]{\mathrm{Rep}(S_{#1})}
\newcommand{\delcatbar}[1]{\underline{\mathrm{Rep}}(S_{#1})}
\newcommand{\undrep}{\underline{\mathsf{Rep}}}
\newcommand{\mfg}{\mathfrak{g}} 
\newcommand{\brak}[1]{\langle #1 \rangle}  



\newcommand{\cC}{{\mathcal C}} 
\newcommand{\cA}{{\mathcal A}}
\newcommand{\cZ}{{\mathcal Z}}
\newcommand{\cD}{{\mathcal D}}
\newcommand{\M}{{\mathcal M}}
\newcommand{\be}{{\bf 1}}
\newcommand{\bl}{{\bf s}}

\newcommand{\eps}{{\varepsilon}}
\newcommand{\sq}{$\square$}
\newcommand{\bi}{\bar \imath}
\newcommand{\bj}{\bar \jmath}
\newcommand{\Ve}{\mbox{Vec}}
\newcommand{\sVec}{\mbox{sVec}}
\newcommand{\Rep}{\mathrm{Rep}}
\newcommand{\Ver}{\mbox{Ver}}
\newcommand{\Mod}{\mbox{Mod}}
\newcommand{\Bimod}{\mbox{Bimod}}
\newcommand{\id}{\mbox{id}}
\newcommand{\inv}{\mbox{inv}}
\newcommand{\ot}{\otimes}
\newcommand{\iHom}{\underline{\mbox{Hom}}}

\def\MS#1{{\color{blue}[MS: #1]}}
\def\MK#1{{\color{red}[MK: #1]}}%

\maketitle
\bibliographystyle{abbrv}

\begin{abstract}
Deep neural networks (DNN) have been widely used and play a major role in the field of  computer vision and autonomous navigation. However, these DNNs are computationally complex and their deployment over resource-constrained platforms is difficult without additional optimizations and customization. 

In this manuscript, we describe an overview of DNN architecture and propose methods to reduce  computational complexity in order to accelerate training and inference speeds to fit them on edge computing platforms with low computational resources.
\end{abstract}


%
%

\section{Introduction} 
\label{section:intro}

Deep neural network design problems support both theoretical and experimental frameworks of machine learning \cite{anthony2009neural,hansen1990neural,hecht1992theory,rowley1998neural,abdi1994neural}. They can detect objects, recognize letters, digits, and symbols, and process scenarios such as running children and a crying woman. They are used in video and monitoring systems to understand established social and power networks via marriage alliance and business relations, to model complex scientific collaborations, to examine multilayered climate dynamics and chain reaction of behavior of atoms in molecular biology. They may be used to represent air transportation networks for various companies, in which multilayer structure depicts flight connections operated by different airline companies. Complex algorithms allow such models to compute and analyze situations, understand scenarios, and react to their surroundings, distinguishing between friendly and hostile environments.

Multilayer machine learning models have become more complex, with millions of  parameters and weights, in order to achieve more accurate and sophisticated outcomes. But all of such operations are not necessary in order to achieve a reasonable output. In fact, they often drain finite computational time and energy resources, not being able to finish the computation within a reasonable time allowed. See, \textit{e.g.}, other research papers that have investigated optimization problems with low tactical computing platforms (\cite{dasari2018complexity,Im-Dasari-quantum-communication-channels, DIB-quantum-comp,IDBS-SWAP-tactical-computing,Im-Dasari-genetic}). 
Therefore, there is an immense problem due to resource constraints on the tactical weights.

The aim of this manuscript is to reduce multilayer machine learning model complexity whilst keep accuracy of the model by at least a reasonable amount, say $(100-\varepsilon)\%$, where $\varepsilon$ is small, \textit{e.g.}, $2\%$. That is, we modify the model in neural networks. The goal is to shrink the architectural aspects of the model so that speed, accuracy, and data are preserved. The unnecessary mathematics and processes are removed, depending on the computation, and we execute this by reducing certain computational aspects of model layers in a systematic way.


\section{Neural network structures}
\label{section:neural-network-structure}

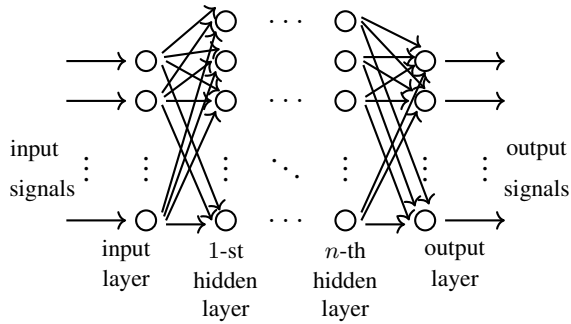
\begin{figure}[ht]
    \centering
\begin{tikzpicture}[scale=0.53]

\node at (-.8,1.75) {\small input};
\node at (-.6,0.75) {\small signals};

\draw[thick,->] (0,4) -- (1.5,4); 
\draw[thick,->] (0,3) -- (1.5,3); 
\draw[thick,->] (0,0) -- (1.5,0); 

\draw[thick,->] (2.4,.2) -- (3.75,4.75);
\draw[thick,->] (2.45,.1) -- (3.75,3.75);
\draw[thick,->] (2.5,0) -- (3.75,2.75);
\draw[thick,->] (2.5,-.1) -- (3.5,-.1);

\draw[thick,->] (2.4,3.2) -- (3.65,5);
\draw[thick,->] (2.45,3.1) -- (3.65,4);
\draw[thick,->] (2.5,3) -- (3.65,3);
\draw[thick,->] (2.4,2.9) -- (3.65,0);

\draw[thick,->] (2.4,4.2) -- (3.75,5.35);
\draw[thick,->] (2.45,4.1) -- (3.85,4.3);
\draw[thick,->] (2.5,4) -- (3.85,3.25);
\draw[thick,->] (2.4,3.9) -- (3.85,0.25);

\draw[thick] (2.25,4) arc (0:360:.25);
\draw[thick] (2.25,3) arc (0:360:.25);
\draw[thick] (2.25,0) arc (0:360:.25);

\node at (0.5,1.5) {\large $\vdots$};
\node at (2,1.5) {\large $\vdots$};
\node at (4,1.5) {\large $\vdots$};
\node at (7,1.5) {\large $\vdots$};
\node at (9,1.5) {\large $\vdots$};
\node at (10.5,1.5) {\large $\vdots$};

\node at (5.5,1.5) {\large $\ddots$};

\draw[thick] (4.25,5) arc (0:360:.25);
\draw[thick] (4.25,4) arc (0:360:.25);
\draw[thick] (4.25,3) arc (0:360:.25);
\draw[thick] (4.25,0) arc (0:360:.25);

\node at (5.5,5) {\large $\ldots$};
\node at (5.5,3) {\large $\ldots$};
\node at (5.5,0) {\large $\ldots$};

\draw[thick] (7.25,5) arc (0:360:.25);
\draw[thick] (7.25,4) arc (0:360:.25);
\draw[thick] (7.25,3) arc (0:360:.25);
\draw[thick] (7.25,0) arc (0:360:.25);

\draw[thick,<-] (9.1,.4) -- (7.4,4.9);
\draw[thick,<-] (8.75,.35) -- (7.45,3.9);
\draw[thick,<-] (8.6,0) -- (7.5,2.9);
\draw[thick,<-] (8.5,-.1) -- (7.5,-.1);

\draw[thick,<-] (9.0,3.3) -- (7.4,5);
\draw[thick,<-] (8.7,3.25) -- (7.45,4);
\draw[thick,<-] (8.6,3) -- (7.5,3);
\draw[thick,<-] (8.8,2.75) -- (7.4,0);

\draw[thick,<-] (8.9,4.35) -- (7.4,5.1);
\draw[thick,<-] (8.75,4.1) -- (7.45,4.1);
\draw[thick,<-] (8.75,3.9) -- (7.5,3.1);
\draw[thick,<-] (9.0,3.73) -- (7.4,0.2);

\draw[thick] (9.25,4) arc (0:360:.25);
\draw[thick] (9.25,3) arc (0:360:.25);
\draw[thick] (9.25,0) arc (0:360:.25);

\draw[thick,->] (9.5,4) -- (11,4); 
\draw[thick,->] (9.5,3) -- (11,3); 
\draw[thick,->] (9.5,0) -- (11,0); 

\node at (1.5,-.75) {\small input}; 
\node at (1.5,-1.5) {\small layer};

\node at (4,-.75) {\small $1$-st}; 
\node at (4,-1.5) {\small hidden};
\node at (4,-2.25) {\small layer};

\node at (7,-.75) {\small $n$-th}; 
\node at (7,-1.5) {\small hidden};
\node at (7,-2.25) {\small layer};

\node at (9.75,-.75) {\small output}; 
\node at (9.75,-1.5) {\small layer};

\node at (11.8,1.75) {\small output};
\node at (11.8,0.75) {\small signals};

\end{tikzpicture}
    \caption{A directed graph of a multilayer machine learning model with connections and weights. It represents a feedforward network with $n+2$ layers, input layer, output layer, and $n$ hidden layers.}
    \label{fig1_01}
\end{figure}

A multilayer machine learning model is an artificial neural network, which is composed of multiple layers of nodes with threshold activation \cite{anthony2009neural,hansen1990neural,hecht1992theory,rowley1998neural,abdi1994neural}. It consists of an input layer, (multiple) hidden layers, and an output layer. The nodes are weighted, with higher weights representing activated neurons. 
See Figure \ref{fig1_01}. 

There are several primary neural network architectural structures. Convolutional neural networks are primarily designed for image recognition, recurrent neural networks best produce predictive results in sequential data, and artificial neural networks model nonlinear problems to predict the output values for given input parameters from their training values. 

\subsection{Convolutional neural networks}
\label{subsection:convolutional-neural-net}
 
In convolutional neural networks, images of a letter, number, or a symbol at a low resolution can be identified by a deep neural network. For example, if the image of a single digit is on a grid of $28$ by $28$ pixels, then the input consists of $28^2$ input data while the output consists of a single number between $0$ to $9$. There are multiple hidden layers, with signals forming a feedforward neural network, flowing only in one direction from input to output, see, \textit{e.g.}, Figure~\ref{fig1_01}.

\subsection{Recurrent neural networks}
\label{subsection:recurrent-neural-net}

\begin{figure}[ht]
    \centering
\begin{tikzpicture}[scale=0.6]
\draw[thick] (0,.3) rectangle (2,7.7);

\draw[thick] (1.35,7) arc (0:360:.35);
\draw[thick] (1.35,6) arc (0:360:.35);
\draw[thick] (1.35,5) arc (0:360:.35);
\draw[thick] (1.35,4) arc (0:360:.35);
\draw[thick] (1.35,3) arc (0:360:.35);
\draw[thick] (1.35,2) arc (0:360:.35);
\draw[thick] (1.35,1) arc (0:360:.35);
\end{tikzpicture}
    \caption{The only structural layer in the entire network.}
    \label{fig1_03}
\end{figure}
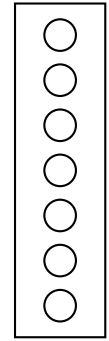

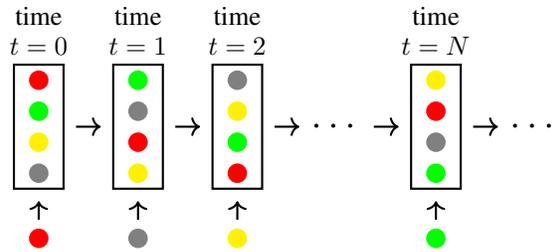
\begin{figure}[ht]
    \centering
\begin{tikzpicture}[scale=0.33]

\node at (1,6.95) {time};
\node at (1,5.75) {$t=0$};

\node at (5,6.95) {time};
\node at (5,5.75) {$t=1$};

\node at (9,6.95) {time};
\node at (9,5.75) {$t=2$};

\node at (17,6.95) {time};
\node at (17,5.75) {$t=N$};

\draw[thick] (0,0) rectangle (2,5);
\draw[thick,->] (2.5,2.5) -- (3.5,2.5);

\draw[thick] (4,0) rectangle (6,5);
\draw[thick,->] (6.5,2.5) -- (7.5,2.5);

\draw[thick] (8,0) rectangle (10,5);
\draw[thick,->] (10.5,2.5) -- (11.5,2.5);

\node at (13,2.5) {\Large $\cdots$};

\draw[thick,->] (14.5,2.5) -- (15.5,2.5);

\draw[thick] (16,0) rectangle (18,5);
\draw[thick,->] (18.5,2.5) -- (19.5,2.5);

\node at (21,2.5) {\Large $\cdots$};

\draw[thick,->] (1,-1.25) -- (1,-.5);
\draw[thick,fill,red] (1.35,-2) arc (0:360:.35);

\draw[thick,->] (5,-1.25) -- (5,-.5);
\draw[thick,fill,gray] (5.35,-2) arc (0:360:.35);

\draw[thick,->] (9,-1.25) -- (9,-.5);
\draw[thick,fill,yellow] (9.35,-2) arc (0:360:.35);

\draw[thick,->] (17,-1.25) -- (17,-.5);
\draw[thick,fill,green] (17.35,-2) arc (0:360:.35);

\draw[thick,fill,red] (1.35,4.375) arc (0:360:.35);
\draw[thick,fill,green] (1.35,3.125) arc (0:360:.35);
\draw[thick,fill,yellow] (1.35,1.875) arc (0:360:.35);
\draw[thick,fill,gray] (1.35,.625) arc (0:360:.35);

\draw[thick,fill,green] (5.35,4.375) arc (0:360:.35);
\draw[thick,fill,gray] (5.35,3.125) arc (0:360:.35);
\draw[thick,fill,red] (5.35,1.875) arc (0:360:.35);
\draw[thick,fill,yellow] (5.35,.625) arc (0:360:.35);

\draw[thick,fill,gray] (9.35,4.375) arc (0:360:.35);
\draw[thick,fill,yellow] (9.35,3.125) arc (0:360:.35);
\draw[thick,fill,green] (9.35,1.875) arc (0:360:.35);
\draw[thick,fill,red] (9.35,.625) arc (0:360:.35);

\draw[thick,fill,yellow] (17.35,4.375) arc (0:360:.35);
\draw[thick,fill,red] (17.35,3.125) arc (0:360:.35);
\draw[thick,fill,gray] (17.35,1.875) arc (0:360:.35);
\draw[thick,fill,green] (17.35,.625) arc (0:360:.35);

\end{tikzpicture}
    \caption{An example of the structure of recurrent neural network, which can be thought of as a passage through time. There are 4 time stamps at $t=0$. At $t=1$, the net takes the output of time $t=0$ and send it back into the net along with the next input. This procedure is repeated in the net. }
    \label{fig1_02}
\end{figure}

If patterns in a data set change with respect to time, then recurrent neural networks would best represent such data set. This machine learning model has a structure with a built-in feedback loop, allowing it to act as a forecasting engine. They are applied from speech recognition to driverless vehicles and robotics. Figure~\ref{fig1_03} shows the one and only structural layer in the entire network. But here, the output of a layer is added to the next input and fed back into the same layer, which is usually the only layer in the entire network. This process could be thought of as a passage through time. 
See Figure~\ref{fig1_02}. Also see  \cite{graves2005framewise,graves2006connectionist,klambauer2017self, lipton2015critical} for more detail. 

Unlike feedforward nets, a recurrent neural network receives a sequence of values as input, as well as produces a sequence of values as output. Its ability to operate with sequences opens up these nets to a wide variety of applications. For example, with a single input and a sequence of outputs, one such application is image captioning. 
On the other hand, a sequence of inputs with a single output may be used for document classification. When both the input and the output are sequences, these nets can classify videos frame by frame. If a time delay is introduced, the neuron network can statistically forecast that demand in supply chain planning.

\subsection{Artificial neural network}
\label{subsection:artificial-neural-net}

Artificial neural networks are computing systems designed to simulate by analyzing and processing information. As the human brain has around $1.0 \times 10^{12}$ neurons, artificial neural networks are designed in programmable machines to behave like interconnected brain cells.

Each connection can transmit a signal to other neurons. An artificial neuron that receives a signal then processes it and can signal neurons connected to it. The signal at a connection is a real number, and the output of each neuron is computed by some nonlinear function of the sum of its inputs.

\section{Neural network learning}
\label{section:neural-network-learning}

Neurons are connected to each other in various patterns, to allow the output of some neurons to become the input of others; they form a directed, weighted graph.

\subsection{Neurons}
\label{subsection:neurons}

Each artificial neuron has inputs and produces a single output which can be sent to multiple other neurons. Initial inputs may be the feature values of some external data, such as images or documents, or they can be the outputs of other neurons. The outputs of the final output neurons of the neural net accomplish the task, such as recognizing an object in an image or letters and symbols.

We follow the following procedure to obtain the output of the neuron. First, we take the weighted sum of all the inputs, weighted by the weights of the connections from the inputs to the neuron. We add a bias term to this sum. This weighted sum is also called the activation. This weighted sum is then passed through a (usually nonlinear) activation function to produce the output. The ultimate outputs accomplish the task.

\subsection{Connections and weights}
\label{subsection:connection-weights}

The edges are called connections. Neurons and connections have a weight that adjusts as learning proceeds. The weight increases or decreases the strength of the signal at a connection. Neurons have a threshold such that a signal is sent only if the aggregate signal crosses that threshold. 

Neurons are aggregated into layers, with different layers performing different transformations on their inputs. Signals travel from the first layer, also known as the input layer, to the last layer, also known as the output layer, possibly after traversing the layers multiple times.

\subsection{Hidden layers}
\label{subsection:hidden-layers}

In multilayer neural networks, hidden layers are located between the input and the output layer of the algorithm. 
The function programmed into the network applies weights to the inputs, adds appropriate bias, and directs them through an activation function as the output.
They perform nonlinear transformations of the inputs entered into the network since they are designed to produce an output specific for an intended result. 
Often times, they are useful when the intended output of the algorithm is a probability, since they take an input and produce an output value $x$, where $0\leq x\leq 1$. 

Hidden layers allow for the function of a neural network to be broken down into specific transformations of the data, where each hidden layer function is specialized to produce a defined output. As an example, a hidden layer that is used to identify ears and eyes cannot solely identify the person. However, when placed in conjunction with additional hidden layers used to identify the facial structure, hair, body type, etc., the neural network can then make predictions and identify the correct individual within visual data.

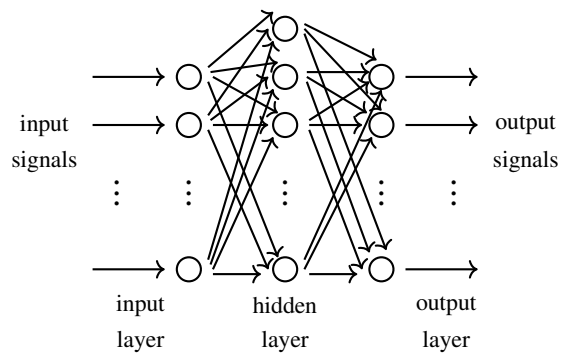
\begin{figure}[ht]
    \centering
    \begin{tikzpicture}[scale=0.64]
    
\node at (-1,3) {\small input};
\node at (-1,2.25) {\small signals};

\draw[thick,->] (0,4) -- (1.5,4); 
\draw[thick,->] (0,3) -- (1.5,3); 
\draw[thick,->] (0,0) -- (1.5,0); 

\draw[thick,->] (2.4,.2) -- (3.75,4.75);
\draw[thick,->] (2.45,.1) -- (3.75,3.75);
\draw[thick,->] (2.5,0) -- (3.75,2.75);
\draw[thick,->] (2.5,-.1) -- (3.5,-.1);

\draw[thick,->] (2.4,3.2) -- (3.65,5);
\draw[thick,->] (2.45,3.1) -- (3.65,4);
\draw[thick,->] (2.5,3) -- (3.65,3);
\draw[thick,->] (2.4,2.9) -- (3.65,0);

\draw[thick,->] (2.4,4.2) -- (3.75,5.35);
\draw[thick,->] (2.45,4.1) -- (3.85,4.3);
\draw[thick,->] (2.5,4) -- (3.85,3.25);
\draw[thick,->] (2.4,3.9) -- (3.85,0.25);

\draw[thick] (2.25,4) arc (0:360:.25);
\draw[thick] (2.25,3) arc (0:360:.25);
\draw[thick] (2.25,0) arc (0:360:.25);

\node at (0.5,1.75) {\Large $\vdots$};
\node at (2,1.75) {\Large $\vdots$};
\node at (4,1.75) {\Large $\vdots$};
\node at (6,1.75) {\Large $\vdots$};
\node at (7.5,1.75) {\Large $\vdots$};

\draw[thick] (4.25,5) arc (0:360:.25);
\draw[thick] (4.25,4) arc (0:360:.25);
\draw[thick] (4.25,3) arc (0:360:.25);
\draw[thick] (4.25,0) arc (0:360:.25);

\draw[thick,<-] (6.1,.4) -- (4.4,4.9);
\draw[thick,<-] (5.75,.35) -- (4.45,3.9);
\draw[thick,<-] (5.6,0) -- (4.5,2.9);
\draw[thick,<-] (5.5,-.1) -- (4.5,-.1);

\draw[thick,<-] (6.0,3.3) -- (4.4,5);
\draw[thick,<-] (5.7,3.25) -- (4.45,4);
\draw[thick,<-] (5.6,3) -- (4.5,3);
\draw[thick,<-] (5.8,2.75) -- (4.4,0);

\draw[thick,<-] (5.9,4.35) -- (4.4,5.1);
\draw[thick,<-] (5.75,4.1) -- (4.45,4.1);
\draw[thick,<-] (5.75,3.9) -- (4.5,3.1);
\draw[thick,<-] (6.0,3.73) -- (4.4,0.2);

\draw[thick] (6.25,4) arc (0:360:.25);
\draw[thick] (6.25,3) arc (0:360:.25);
\draw[thick] (6.25,0) arc (0:360:.25);

\draw[thick,->] (6.5,4) -- (8,4);
\draw[thick,->] (6.5,3) -- (8,3);
\draw[thick,->] (6.5,0) -- (8,0);

\node at (1,-.75) {\small input};
\node at (1,-1.5) {\small layer};

\node at (4,-.75) {\small hidden};
\node at (4,-1.5) {\small layer};

\node at (7.35,-.75) {\small output};
\node at (7.35,-1.5) {\small layer};

\node at (9,3) {\small output};
\node at (9,2.25) {\small signals};
    
    \end{tikzpicture}
    \caption{A simple neural network consists of exactly one hidden layer.}
    \label{fig1_05}
\end{figure}

Figure~\ref{fig1_05} is an example of a simple neural network, which has one hidden layer, while Figure~\ref{fig1_01} is an example of a more general neural network, with $n\geq 1$ hidden layers. 

\section{Propagation and backpropagation}
\label{subsection:propagation-backpropagation}

The problem of vanishing gradient is worse for recurrent neural networks. This is because each time step is the equivalent of an entire layer in a feedforward network. So training a recurrent neural network for $10,000$ time steps is equivalent to training a $10,000$-layer feedforward net, which leads to exponentially small gradients and a decay of information through time. 

One way to address this problem is by the method of gating, which is a technique to help decide when to forget the current input, and when to remember it for future time steps.  The most popular gating types are gated recurrent unit (GRU) and long short-term memory (LSTM), while other techniques include gradient clipping, steeper gates, and better optimizers.

When it comes to training a recurrent network, graphics processing units (GPUs) are preferred over central processing units (CPUs), via accumulated evidence on text processing tasks like sentiment analysis and helpfulness extraction. 
GPUs train the neural network 250 times faster, where GPUs can complete a computation in $1$ day versus over $8$ months using CPUs.

\begin{figure}[ht]
    \centering
\begin{tikzpicture}[scale=0.3]

\node at (1,13.95) {time};
\node at (1,12.75) {$t=0$};

\node at (5,13.95) {time};
\node at (5,12.75) {$t=1$};

\node at (9,13.95) {time};
\node at (9,12.75) {$t=2$};

\node at (17,13.95) {time};
\node at (17,12.75) {$t=N$};

\node at (13,6) {\Large $\ddots$};

\draw[thick,->] (1,5.5) -- (1,6.25);
\draw[thick,->] (5,5.5) -- (5,6.25);
\draw[thick,->] (9,5.5) -- (9,6.25);
\draw[thick,->] (17,5.5) -- (17,6.25);

\draw[thick] (0,0) rectangle (2,5);
\draw[thick,->] (2.5,2.5) -- (3.5,2.5);

\draw[thick] (4,0) rectangle (6,5);
\draw[thick,->] (6.5,2.5) -- (7.5,2.5);

\draw[thick] (8,0) rectangle (10,5);
\draw[thick,->] (10.5,2.5) -- (11.5,2.5);

\node at (13,2.5) {\Large $\cdots$};

\draw[thick,->] (14.5,2.5) -- (15.5,2.5);

\draw[thick] (16,0) rectangle (18,5);
\draw[thick,->] (18.5,2.5) -- (19.5,2.5);

\node at (21,2.5) {\Large $\cdots$};

\draw[thick,->] (1,-1.25) -- (1,-.5);
\draw[thick,fill,red] (1.35,-2) arc (0:360:.35);

\draw[thick,->] (5,-1.25) -- (5,-.5);
\draw[thick,fill,gray] (5.35,-2) arc (0:360:.35);

\draw[thick,->] (9,-1.25) -- (9,-.5);
\draw[thick,fill,yellow] (9.35,-2) arc (0:360:.35);

\draw[thick,->] (17,-1.25) -- (17,-.5);
\draw[thick,fill,green] (17.35,-2) arc (0:360:.35);

\draw[thick,fill,red] (1.35,4.375) arc (0:360:.35);
\draw[thick,fill,green] (1.35,3.125) arc (0:360:.35);
\draw[thick,fill,yellow] (1.35,1.875) arc (0:360:.35);
\draw[thick,fill,gray] (1.35,.625) arc (0:360:.35);

\draw[thick,fill,green] (5.35,4.375) arc (0:360:.35);
\draw[thick,fill,gray] (5.35,3.125) arc (0:360:.35);
\draw[thick,fill,red] (5.35,1.875) arc (0:360:.35);
\draw[thick,fill,yellow] (5.35,.625) arc (0:360:.35);

\draw[thick,fill,gray] (9.35,4.375) arc (0:360:.35);
\draw[thick,fill,yellow] (9.35,3.125) arc (0:360:.35);
\draw[thick,fill,green] (9.35,1.875) arc (0:360:.35);
\draw[thick,fill,red] (9.35,.625) arc (0:360:.35);

\draw[thick,fill,yellow] (17.35,4.375) arc (0:360:.35);
\draw[thick,fill,red] (17.35,3.125) arc (0:360:.35);
\draw[thick,fill,gray] (17.35,1.875) arc (0:360:.35);
\draw[thick,fill,green] (17.35,.625) arc (0:360:.35);

\draw[thick] (0,7) rectangle (2,12);
\draw[thick,->] (2.5,9.5) -- (3.5,9.5);

\draw[thick] (4,7) rectangle (6,12);
\draw[thick,->] (6.5,9.5) -- (7.5,9.5);

\draw[thick] (8,7) rectangle (10,12);
\draw[thick,->] (10.5,9.5) -- (11.5,9.5);

\node at (13,9.5) {\Large $\cdots$};

\draw[thick,->] (14.5,9.5) -- (15.5,9.5);

\draw[thick] (16,7) rectangle (18,12);
\draw[thick,->] (18.5,9.5) -- (19.5,9.5);

\node at (21,9.5) {\Large $\cdots$};

\draw[thick,fill,gray] (1.35,11.375) arc (0:360:.35);
\draw[thick,fill,green] (1.35,10.125) arc (0:360:.35);
\draw[thick,fill,red] (1.35,8.875) arc (0:360:.35);
\draw[thick,fill,yellow] (1.35,7.625) arc (0:360:.35);

\draw[thick,fill,green] (5.35,11.375) arc (0:360:.35);
\draw[thick,fill,gray] (5.35,10.125) arc (0:360:.35);
\draw[thick,fill,yellow] (5.35,8.875) arc (0:360:.35);
\draw[thick,fill,red] (5.35,7.625) arc (0:360:.35);

\draw[thick,fill,yellow] (9.35,11.375) arc (0:360:.35);
\draw[thick,fill,red] (9.35,10.125) arc (0:360:.35);
\draw[thick,fill,gray] (9.35,8.875) arc (0:360:.35);
\draw[thick,fill,green] (9.35,7.625) arc (0:360:.35);

\draw[thick,fill,red] (17.35,11.375) arc (0:360:.35);
\draw[thick,fill,yellow] (17.35,10.125) arc (0:360:.35);
\draw[thick,fill,green] (17.35,8.875) arc (0:360:.35);
\draw[thick,fill,gray] (17.35,7.625) arc (0:360:.35);

\end{tikzpicture}
    \caption{An example of stacking in recurrent neural network. }
    \label{fig1_04}
\end{figure}
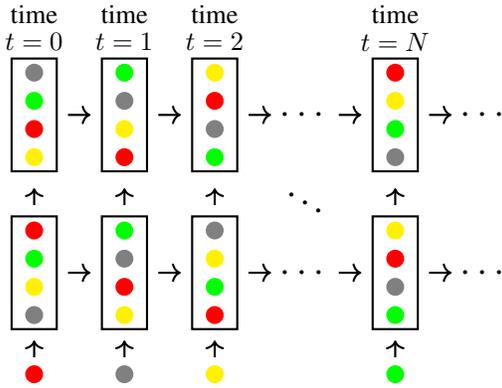

A notion called stacking (see Figure~\ref{fig1_04})  consists of a more complex output than processing a single recurrent neural network.

\section{The structure of neural network models} 
\label{section:structure-neural-network-models}

A network is a graph that represents a complex system, which uses linear and nonlinear algebraic operations, statistics, linear algebra, and any other necessary mathematical techniques to obtain accurate outcomes.

\begin{figure}[ht]
    \centering
\begin{tikzpicture}[scale=0.5]

\draw[thick] (0,6) arc (0:360:1);
\draw[thick] (0,3.5) arc (0:360:1);
\node at (-1,2) {\large $\vdots$};
\draw[thick] (0,0) arc (0:360:1);

\draw[thick] (6,6) arc (0:360:1);

\draw[thick,->] (.5,0) -- (4.6,4.6);
\draw[thick,->] (.5,3.5) -- (4,5);
\draw[thick,->] (.5,6) -- (3.5,6);

\node at (-1,6) {\large $a_1$};
\node at (-1,3.5) {\large $a_2$};
\node at (-1,0) {\large $a_{n}$};

\node at (2,6.5) {\large $w_1$};
\node at (2,4.8) {\large $w_2$};
\node at (2,2.8) {\large $w_n$};

\end{tikzpicture}
    \caption{An instance of the input layer activations all connected to a node in the second layer.
    }
    \label{fig1_06}
\end{figure}
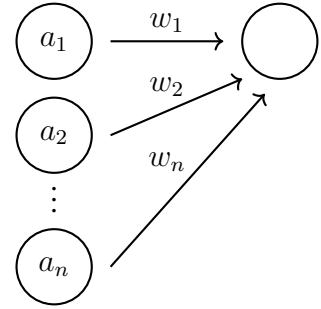

Assign a weight $w_i$ to each edge (connection) between neuron from the first layer to our neuron, where $w_i\in \mathbb{R}$ is allowed to be any positive or negative number. 
Take all those activations $a_i$ from the input (first) layer and compute their weighted sum: 
\begin{equation}
\label{eqn:weighted-sum-activation}
\sum_{i=1}^n w_ia_i = w_1a_1 + w_2a_2 + \ldots + w_n a_n.
\end{equation} 
See Figure~\ref{fig1_06}. 

Since the weighted sum 
$\sum_{i=1}^n w_ia_i$ is allowed to be any value, activation values should be between $0$ and $1$ in order to make sense, \textit{e.g.}, for an image processing scenario. So in such example, normalize the weighted sum into the range  $(0,1)$ using the logistic curve  
\[  
\sigma: \mathbb{R}\rightarrow (0,1), 
\mbox{ where }
\sigma(x) = \frac{1}{1+e^{-x}}. 
\]  
The activation $\sigma(\sum_{i=1}^n w_i a_i)$ of the neuron is essentially a measure of the positivity of the relevant weighted sum. 
If the activation is meaningful only when the weighted sum is bigger than a bias $b$, then  modify the activation by subtracting $b$:  $\sigma(\sum_{i=1}^n w_i a_i - b)$. 
So the weights give what pixel pattern this neuron in the second layer is picking up on and the bias says how high the weighted sum needs to be before the neuron begins to be meaningfully active.

The above analysis is for a single neuron, so every neuron in the second layer is connected to all input neurons from the first layer, and each of those connections has its own weight (on the edge) and bias associated with it. 
This is just the connections from the first layer to the second layer. Connections between all the other layers have their own weights and biases associated with them. 

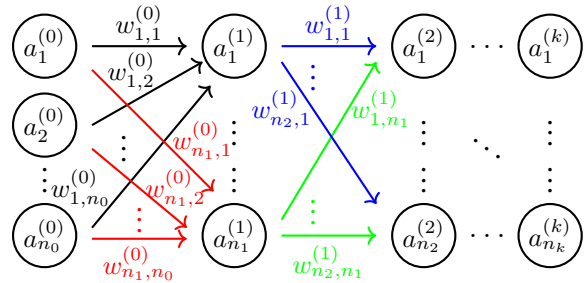
\begin{figure}[ht]
    \centering
\begin{tikzpicture}[scale=0.42]

\draw[thick] (0,6) arc (0:360:1);
\draw[thick] (0,3.5) arc (0:360:1);
\node at (-1,2) {\large $\vdots$};
\draw[thick] (0,0) arc (0:360:1);

\draw[thick] (6,6) arc (0:360:1);
\node at (5,3.5) {\large $\vdots$};
\node at (5,2) {\large $\vdots$};
\draw[thick] (6,0) arc (0:360:1);

\draw[thick,->] (.5,0.25) -- (4.25,4.75);
\node at (1.5,3) {\large $\vdots$};
\draw[thick,->] (.5,3.5) -- (4,5.5);
\draw[thick,->] (.5,6) -- (3.5,6);

\node at (-1,6) {$a_1^{(0)}$};
\node at (-1,3.5) {$a_2^{(0)}$};
\node at (-1,0) {$a_{n_0}^{(0)}$};

\node at (2,6.7) {$w_{1,1}^{(0)}$};
\node at (1.75,5.2) {$w_{1,2}^{(0)}$};
\node at (.15,1.5) {$w_{1,n_0}^{(0)}$};

\draw[thick,red,->] (.5,5.25) -- (4.35,1.4);
\draw[thick,red,->] (.5,2.75) -- (3.5,.25);
\node[red] at (2,.75) {\large $\vdots$};
\draw[thick,red,->] (.5,-.1) -- (3.5,-.1);

\draw[thick] (12,6) arc (0:360:1);
\node at (11,3.5) {\large $\vdots$};
\node at (11,2) {\large $\vdots$};
\draw[thick] (12,0) arc (0:360:1);

\node at (5,6) {$a_1^{(1)}$};
\node at (5,0) {$a_{n_1}^{(1)}$};

\node[red] at (4,3) {$w_{n_1,1}^{(0)}$};
\node[red] at (3.15,1.5) {$w_{n_1,2}^{(0)}$};
\node[red] at (2,-.9) {$w_{n_1,n_0}^{(0)}$};

\draw[thick,->,green] (6.5,.5) -- (9.5,5.5);
\node[green] at (7.5,1) {\large $\vdots$};
\draw[thick,->,green] (6.5,0) -- (9.5,0);

\node[green] at (9.65,4) {$w_{1,n_1}^{(1)}$};
\node[green] at (8,-.85) {$w_{n_2,n_1}^{(1)}$};

\node at (11,6) {$a_1^{(2)}$};
\node at (11,0) {$a_{n_2}^{(2)}$};

\node[blue] at (8,6.7) {$w_{1,1}^{(1)}$};
\node[blue] at (6.5,4) {$w_{n_2,1}^{(1)}$};

\draw[thick,->,blue] (6.5,6) -- (9.5,6); 
\node[blue] at (7.5,5.25) {\large $\vdots$};
\draw[thick,->,blue] (6.5,5.5) -- (9.5,1); 

\node at (13,6) {\large $\ldots$};
\node at (13,3) {\large $\ddots$};
\node at (13,0) {\large $\ldots$};

\draw[thick] (16,6) arc (0:360:1);
\node at (15,3.5) {\large $\vdots$};
\node at (15,2) {\large $\vdots$};
\draw[thick] (16,0) arc (0:360:1);

\node at (15,6) {$a_1^{(k)}$};
\node at (15,0) {$a_{n_k}^{(k)}$};

\end{tikzpicture}
    \caption{General neural network. }
    \label{fig1_07}
\end{figure}

Representing this mathematically, 
consider the more general case in Figure~\ref{fig1_07}.
Here, we have 
\begin{equation}
\label{eqn:weighted-sum-activation-general}
\begin{aligned}
a_1^{(1)} 
&= \sigma\left(w_{1,1}^{(0)}a_1^{(0)}
+\ldots + w_{1,n_0}^{(0)}a_{n_0}^{(0)} + b_1^{(0)}\right) \\ 
a_2^{(1)} 
&= \sigma\left(w_{2,1}^{(0)}a_1^{(0)}+ \ldots + 
w_{2,n_0}^{(0)}a_{n_0}^{(0)} + b_2^{(0)}\right) \\ 
\vdots &\qquad \qquad \qquad \vdots \\ 
a_{n_1}^{(1)} 
&= \sigma\left(w_{n_1,1}^{(0)}a_1^{(0)}+\ldots + 
w_{n_1,n_0}^{(0)}a_{n_0}^{(0)} + b_{n_1}^{(0)}\right) \\ 
\vdots &\qquad \qquad \qquad \vdots \\ 
\end{aligned}
\end{equation}
Rewrite \eqref{eqn:weighted-sum-activation-general} more compactly as matrices, $\textit{i.e.}$,
\begin{align*}
\tiny 
\begin{pmatrix}
a_1^{(1)} \\ 
a_2^{(1)} \\ 
\vdots  \\ 
a_{n_1}^{(1)} \\ 
\end{pmatrix} 
&=    
\widetilde{\sigma} \left( 
\begin{pmatrix}
w_{1,1}^{(0)}& \ldots & 
w_{1,n_0}^{(0)} \\ 
w_{2,1}^{(0)}& \ldots & 
w_{2,n_0}^{(0)} \\ 
& & & \\ 
w_{n_1,1}^{(0)}& \ldots &  
w_{n_1,n_0}^{(0)} \\ 
\end{pmatrix}
\begin{pmatrix}
a_1^{(0)} \\ 
a_2^{(0)} \\ 
\vdots  \\ 
a_{n_0}^{(0)} \\ 
\end{pmatrix} \right. \\ 
&\hspace{2cm} + 
\left.
\begin{pmatrix}
b_1^{(0)} \\ 
b_2^{(0)} \\ 
\vdots  \\ 
b_{n_1}^{(0)} \\ 
\end{pmatrix} 
\right), 
\end{align*} 
or  
$a^{(1)} = \widetilde{\sigma} (W^{(0)}a^{(0)}+b^{(0)})$, where 
\begin{align*}
\widetilde{\sigma}
\left(\begin{pmatrix}
v_1 \\ 
v_2 \\ 
\vdots \\ 
v_n \\ 
\end{pmatrix}\right) &:= 
\begin{pmatrix}
\sigma(v_1) \\ 
\sigma(v_2) \\ 
\vdots \\ 
\sigma(v_n) \\ 
\end{pmatrix}, \\
a^{(i)} = 
\begin{pmatrix}
a_1^{(i)} \\ 
a_2^{(i)} \\ 
\vdots \\ 
a_{n_i}^{(i)} \\ 
\end{pmatrix}, 
\:\: 
&\mbox{ and } 
\:\:  
b^{(i)} = 
\begin{pmatrix}
b_1^{(i)} \\ 
b_2^{(i)} \\ 
\vdots \\ 
b_{n_{i+1}}^{(i)} \\ 
\end{pmatrix},  
\end{align*}
etc. 
Figure~\ref{fig1_07} has $\wt+\biases$ parameters, where 
\begin{align*}
\wt &= a_{n_0}^{(0)} a_{n_1}^{(1)} + 
a_{n_1}^{(1)} a_{n_2}^{(2)} + 
\ldots 
+ 
a_{n_{k-1}}^{(k-1)} a_{n_k}^{(k)} 
 \\ 
&\:\:\:\:\mbox{ and }\:\:\:\:  \biases = \sum_{i=1}^k a_{n_i}^{(i)}, 
\end{align*}
where $\wt$ is the number of weights and $\biases$ is the number of biases.

So for machine learning, we modify the computer with a valid setting for all of these weights and biases so that it will solve the original problem. 
It seems plausible to change the structure or improve the neural network in order to make the model more efficient. But if the network does work, and not for the reasons that we may expect, then investigating what the weights and biases are doing is an adequate way to challenge one's assumptions and explore the full spectrum of possible solutions. 

As we move from the first layer of edges to the second layer of edges in the general neural network, see Figure~\ref{fig1_07}, we are applying matrix and column vector products and a logistic curve again, analogous to \eqref{eqn:weighted-sum-activation-general}. A specific number that neurons hold depends on the input that has been fed into the model. So each neuron should be thought of as a function. 


\section{Backpropagation algorithm}
\label{section:backpropagation}
Consider the general multilayered neural network in Figure~\ref{fig1_07}.
Let 
$C_k = \displaystyle{\sum_{j=1}^{n_k}} (a_j^{(k)}-y_j^{(k)})^2$ be the cost function in the output layer, where $y_i^{(k)}$'s are the desired output. 

Let $z_j^{(k)} = w_{j,1}^{(k-1)}a_1^{(k-1)}+\ldots + w_{j,n_{k-1}}^{(k-1)}a_{n_{k-1}}^{(k-1)}+b_j^{(k-1)}$, where $1\leq j\leq n_k$.  
Let $a_{j}^{(k)}=\sigma_k(z_j^{(k)})$, where $\sigma_k$ is a nonlinear function. 
The derivative of the cost function with respect to a weight is 
\begin{equation}
    \label{eqn:change-cost-weight-k}
    \begin{split}
    \frac{\partial C_k}{\partial w_{j,u}^{(k-1)}} 
    &= 
    \frac{\partial z_j^{(k)}}{\partial w_{j,u}^{(k-1)}}
    \frac{\partial a_j^{(k)}}{\partial z_j^{(k)}}
    \frac{\partial C_k}{\partial a_j^{(k)}} \\ 
    &= a_u^{(k-1)}\sigma_k'
    (z_j^{(k)}) 
    \left(2 \left(a_j^{(k)} - y_j^{(k)} \right) \right), 
     \end{split} 
\end{equation} 
where 
    $1\leq j\leq n_k$ and 
    $1\leq u\leq n_{k-1}$.

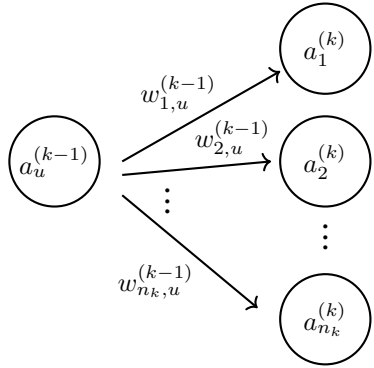
\begin{figure}[ht]
    \centering
\begin{tikzpicture}[scale=0.6]

\draw[thick] (0,3.5) arc (0:360:1);

\draw[thick] (6,6) arc (0:360:1);
\draw[thick] (6,3.5) arc (0:360:1);
\node at (5,2) {\Large $\vdots$};
\draw[thick] (6,0) arc (0:360:1);

\node at (1.5,2.8) {\Large $\vdots$};
\draw[thick,->] (.5,3.5) -- (4,5.5);

\node at (-1,3.5) {$a_u^{(k-1)}$};

\node at (1.75,5) {$w_{1,u}^{(k-1)}$};
\node at (2.95,4) {$w_{2,u}^{(k-1)}$};

\draw[thick,->] (.5,2.75) -- (3.5,.25);
\draw[thick,->] (.5,3.2) -- (3.8,3.5);

\node at (5,6) {$a_1^{(k)}$};
\node at (5,3.5) {$a_2^{(k)}$};
\node at (5,0) {$a_{n_k}^{(k)}$};

\node at (1.25,.85) {$w_{n_k,u}^{(k-1)}$};
 
\end{tikzpicture}
    \caption{The activation in the previous layer influences the cost function through multiple paths.}
    \label{fig1_09}
\end{figure}

The derivative of the cost with respect to one of the activations in layer $k-1$ is 
\begin{equation}
    \label{eqn:change-in-cost-neuron}
\frac{\partial C_k}{\partial a_u^{(k-1)}} = \sum_{j=1}^{n_k} \frac{\partial z_j^{(k)}}{\partial a_u^{(k-1)}}
\frac{\partial a_j^{(k)}}{\partial z_j^{(k)}}
\frac{\partial C_k}{\partial a_j^{(k)}}, 
\end{equation}
where $1\leq u\leq n_{k-1}$.  
Here, a neuron in layer $k-1$ influences the cost function through multiple paths, so we sum over activations in layer $k$. 
This tells us how sensitive the cost function is relative to the activation of the previous layer (see Figure~\ref{fig1_09}).

Finally, we have 
\begin{equation}
    \label{eqn:change-cost-change-bias}
    \frac{\partial C_k}{\partial b_j^{(k)}} 
    = 
    \frac{\partial z_j^{(k)}}{\partial b_j^{(k)}}
    \frac{\partial a_j^{(k)}}{\partial z_j^{(k)}}
    \frac{\partial C_k}{\partial a_j^{(k)}}
    = \sigma_k'(z_j^{(k)})2(a_j^{(k)}-y_j^{(k)}).  
\end{equation}
More generally, 
\begin{equation}
    \label{eqn:change-cost-weight-general}
    \begin{split}
    \frac{\partial C_l}{\partial w_{j,u}^{(l-1)}} 
    &= 
    \frac{\partial z_j^{(l)}}{\partial w_{j,u}^{(l-1)}}
    \frac{\partial a_j^{(l)}}{\partial z_j^{(l)}}
    \frac{\partial C_l}{\partial a_j^{(l)}} \\ 
     &= a_u^{(l-1)}\sigma_{l}'(z_j^{(l)}) 2(a_j^{(l)} -y_j^{(l)} ),  \\ 
    \end{split}     
\end{equation}
where $1\leq l\leq k$. 

Thus, the derivative of the full cost function 
\[
\frac{\partial C}{\partial w^{(u)}}
= \frac{1}{N} 
\sum_{v=1}^{N} 
\frac{\partial C^{(v)}}{\partial w^{(u)}}
\] 
is the average of all training examples, and 
\[ 
\nabla C = 
\begin{pmatrix}
\frac{\partial C}{\partial w_{1,1}^{(0)}} \\ 
\vdots \\ 
\frac{\partial C}{\partial b_{n_1}^{(0)}} \\ 
\vdots \\ 
\frac{\partial C}{\partial w_{1,1}^{(k)}} \\ 
\vdots \\ 
\frac{\partial C}{\partial b_{n_k}^{(k)}} \\ 
\end{pmatrix}. 
\]

\section{Optimization and neural network reduction}
\label{section:network-reduction}
Optimization on tactical computing platforms are very important in today's society  (\cite{dasari2018complexity,Im-Dasari-quantum-communication-channels, DIB-quantum-comp,IDBS-SWAP-tactical-computing,Im-Dasari-genetic}), as more advanced models become quickly developed but computational resources remain limited. This prevents even the most advanced computing platforms to complete many of their algorithms.

Optimizers define how neural networks learn; they find the values of parameters/predictors such that a loss function is minimized. In general, the optimizers do not know the terrain of the loss so they need to minimize the function essentially blindfolded. 

The original optimizer for deep neural networks involves taking small steps iteratively until the correct weights are reached. However, the problem is the weights are updated once after seeing the entire data set. So this gradient is typically large, and the weights could lead to larger jumps. It may also hover over its optimal value without actually being able to reach it. The solution to this is to update the parameters more frequently. 

One example is to use stochastic gradient descent, which updates the weights after seeing each data point, instead of the entire data set. But this may make exceedingly noisy jumps that move away from the optimal values since it is influenced by \textit{every} single sample. Because of this, mini-batch gradient descent is used as a compromise, updating the parameters after \textit{several} samples. 

Another way to decrease the noise of stochastic gradient descent is to add the concept of momentum. The parameters of a model may have the tendency to change in one direction, typically if examples follow a similar pattern. With this momentum, the model can learn faster by paying little attention to the few examples that throw it off from time-to-time. But there is a problem here also. The jumps could be too large to the point that the cost function actually moves away from the minimum, optimal values. That is, choosing to blindly ignore samples simply because they are not typical may be a costly mistake, resulting in a loss. However, adding an acceleration term will help.  The weights for the model-in-training (neural network edge values that are being adjusted) could become larger, with little to no effect from the outliers.
This would show that discarding the outliers would not lead to a drastic loss in order to fine-tune the current model.

With multiple predictors, the learning rate is fixed for every parameter, but one can impose adaptive learning rate to each parameter, where different size step is taken for every parameter. Adaptive learning rate optimizers are able to learn more along one direction than another, so they can traverse certain types of terrain. 
Furthermore, momentum updates for each parameter could also be imposed.

The mathematics behind neural networks reduces to using calculus and other analytical techniques to find the minimum of the cost function. Conceptually, we are thinking of each neuron as being connected to all the neurons in the previous layer. The weights $w_i$ or $w_{i,j}^{(k)}$ in the weighted sum defining its activation in \eqref{eqn:weighted-sum-activation} and \eqref{eqn:weighted-sum-activation-general} are the strengths of those connections. The bias $b$ is an indication of whether or not that neuron tends to be active or inactive. 

One constructs a cost function to determine wrong or erroneous output, i.e., it is the sum of the square of the differences between actual and predicted output activations. That is, a cost function $C$ has $\wt+\biases$ as it input and a single number as its output, which says how bad its weights and biases are, and the way that it is defined depends on the network's behavior over all the thousands of pieces of labelled training data, which is the cost of a single training example. 

We want to minimize the cost function by changing all of the weights and biases (all connections) to most efficiently decrease the cost. But the smallest local value of the cost function is not necessarily the global minimum value. In fact, attaining the global minimum may be computationally heavy, possibly wasting time and resources, since we are working with possibly millions of weights and biases. One uses gradient descent 
$-\nabla C(w_{1,1}^{(0)},\ldots, w_{n_1,n_0}^{(0)},
\ldots , w_{1,1}^{(k)},\ldots, w_{n_1,n_0}^{(k)}$,  $b_1^{(k)},\ldots, b_{n_{k-1}}^{(k)}
)$, 
which is the direction of the steepest descent, \textit{i.e.}, it maximally decreases the output of the cost function as quickly as possible.
For example, see Figure~\ref{fig1_08}. 

\begin{figure}[ht]
    \centering
\begin{tikzpicture}[scale=0.4]
\node at (0,0) {$-\nabla C
\left(
\begin{pmatrix}
w_{1,1}^{(0)} \\ 
\vdots \\ 
w_{n_1,n_0}^{(0)} \\ 
\vdots \\ 
w_{1,1}^{(k)} \\ 
\vdots \\ 
w_{n_1,n_0}^{(k)} \\ 
\end{pmatrix}^t
  \right) =
\begin{pmatrix}
0.25\\ 
\vdots \\ 
-0.05 \\ 
\vdots \\ 
1.034 \\ 
\vdots \\ 
-2.98 \\ 
\end{pmatrix}$}; 

\node at (9.3, 4.20) {\small $\uparrow w_{1,1}^{(0)}$};

\node at (9.3,2.75) {\large $\vdots$};

\node at (9.3, 1.5) {\small $\downarrow w_{n_1,n_0}^{(0)}$ by $\approx 0$};

\node at (9.3,0.25) {\large $\vdots$};

\node at (9.3,-1.2) {\small $\uparrow w_{1,1}^{(k)}$  by a lot};

\node at (9.3,-2.5) {\large $\vdots$};

\node at (9.3,-3.9) {\small $\downarrow w_{n_1,n_0}^{(k)}$ by a lot};

\end{tikzpicture}
    \caption{The notation $t$ is the transpose of the column vector. This is an example of the negative of the gradient function, paving a way for a strategy on a neural network to minimize the cost function. The relative magnitude of the components tells us how sensitive the cost function is to each weight and bias, \textit{i.e.}, which changes matter more.}
    \label{fig1_08}
\end{figure}

We see in Figure~\ref{fig1_08} that an adjustment to some of the weights have a greater impact on the cost function than the adjustment of other weights. Thus, some of the connections matter much more for the training data.  The gradient descent function encodes the relative importance of each weight and bias. When changing by some small multiple of $-\nabla C$, the relative importance of the weights and bias remain the same. 
As we can see, this technique is computationally heavy due to immense set of input data. 
Furthermore, the testing data may show that the number of correct outputs over the total number of simulations may be close to approximately $95-98\%$ of testing accuracy, not $100\%$, as we would like. 

Backpropagation is an algorithm for computing the gradient descent. Theoretically, the way we adjust weights and biases for a single gradient descent step also depends on every single training example (since each step uses every example), but for computational efficiency, we will keep from needing to obtain every single example for every single step. 

First, consider one single example, for example, the image of the number $8$. We will discuss the effects this one training example have on how the weights and biases get adjusted. 
If the neural network is not yet well-trained, activations in the output will appear random. For example, there are 10 outputs labelled from $0$ to $9$ and we may have 
\begin{align*}
    a_1^{(k)} &= 0.4 \\ 
    a_2^{(k)} &= 0.7 \\ 
    a_3^{(k)} &= 0.2 \\ 
    a_4^{(k)} &= 0.1 \\     
    a_5^{(k)} &= 0.0 \\ 
    a_6^{(k)} &= 0.4 \\ 
    a_7^{(k)} &= 1.0 \\ 
    a_8^{(k)} &= 0.1 \\ 
    a_9^{(k)} &= 0.0 \\ 
    a_{10}^{(k)} &= 0.3 
    \end{align*}
where $a_{j}^{(k)}$ is the neural network output for the number $j-1$, $1\leq j\leq 10$. If the network is well-trained, then all $a_{j}^{(k)}$ should go down to $0$, except $a_9^{(k)}$ since we are considering the image of the number $8$, so $a_9^{(k)}$ should go up to $1$. We cannot change these activations but instead, we may influence weights and biases. Moreover, the sizes of these nudges are proportional to how far away each current value is from its target value. For example, the increase to that number $8$ neuron activation, \textit{i.e.}, $a_9^{(k)}$, is more important than the decrease to the number $3$ neuron, \textit{i.e.}, $a_{4}^{(k)}$, which is fairly close to where it should be. 

Focusing on $a_{9}^{(k)}$, this activation $a_{9}^{(k)}$  is defined as the weighted sum of all of the activations in the previous layer, plus a bias, and then it has been substituted into a nonlinear function $\sigma$, \textit{cf.}, \eqref{eqn:weighted-sum-activation-general}. We see that there are three different ways to increase the activation: increase the bias, increase the weights, or change the activations from the previous layer. 

Consider the weights. In order to adjust the weights, notice that the weights have differing levels of influence. The connections with the brightest neurons from the preceding layer have the biggest effect since those weights are multiplied by larger activation values. If we increase one of those weights, it has a stronger influence on the ultimate cost function than increasing the weights of connections with dimmer neurons. 

Note that when we consider the gradient descent, we care not only the sign of each component, but we care about the magnitude of each component. That is, the neurons that are firing when seeing the number $8$ get even more greatly linked to those firing when thinking about the number $8$. 

The last way to increase this neuron's activation is by changing all the activations from the previous layer, and in proportion to each associated weight. Namely, if everything connected to that digit $8$ neuron with a positive weight got brighter and everything connected with a negative weight got dimmer, then the digit $8$ neuron would become more active. 
We do not have direct influence on these activations, but we only have control over the weights and biases. 

Now, focusing on all the other neurons $a_j^{(k)}$ in the output layer, where $j\not=9$, we want all of the other neurons in the last layer to become less active; each of those other output neurons has its own algorithm about what should happen to that second-to-last layer. 
We thus need to put together the strategy for the output neuron $8$ along with the strategies for all the other output neurons for what should happen to this second-to-last (hidden) layer,  proportionately to the weights and to how much each of those neurons needs to change. 

Applying backpropagation is adding together these desired effects to obtain a list of nudges that we want to occur in the second-to-last layer. After we apply this, we recursively apply this algorithm to the relevant weights and biases that determine those values by repeating this process and moving backwards through the network. 

This is how a single training example wishes to nudge each one of those weights and biases. If we only focused on one simple example of $8$, then the network would ultimately be incentivized just to classify all images as an $8$. So what we need to do is go through this same backpropagation routine for every other training example, recording how each of them would like to change the weights and the biases. Then we average together these desired changes, over all training data. 
This collection of the average changes to each weight and bias is the negative gradient of the cost function. 

Gradient descent in practice takes large amount of computational time to add up the influence of every single training example, every single gradient descent step. 
So what we do instead is stochastic gradient descent, \textit{i.e.}, 
we randomly shuffle our training data and then divide it into groups of mini-batches, for example, each one having 1000 training examples. Then we compute a gradient descent step using backpropagation according to the mini-batch. 
This doesn't give the actual gradient descent of the cost function, which depends on all of the training data. So this is not the most efficient step downhill, but we do obtain a fairly decent approximation, and it gives us a significant computational speed up. That is, if we plot the trajectory of the network under the relevant cost surface, it would be similar to a person aimlessly walking downhill, taking quick steps, rather than a carefully calculating person determining the exact downhill direction of each infinitesimal step, before taking a very slow and careful step in that direction.

\section{Neural network complexity}
\label{section:complexity}
Numerous labelled training data are needed, like handwritten numbers, letters, and symbols, or people labelling tens of thousands of images. 
Neural networks are governed by bandwidth and performance. As there is a higher demand for artificial intelligence, there is a greater need for bandwidth reduction and performance bump, which will allow one to load less data from system memory into the local memory and overall from the system. 

Although researchers have investigated many ways towards neural network reduction, we focus on certain convolution called Rectified Linear Units (ReLU) or rectified linear activation function. 
That is, we insert a new layer to create more sparsity in the weights and in the activations. ReLU is a piecewise linear function whose output will be the input directly if it is positive, or else, its output is zero.  That is, it's the linear function 
\[ 
f(x) = 
\begin{cases} 
0 &\mbox{ if } x \leq 0, \\ 
x &\mbox{ if } x > 0. \\
\end{cases} 
\] 
However, for our manuscript, we modify the function as 
\begin{equation}
\label{eqn:modified-ReLU}
f(x) = 
\begin{cases} 
0 &\mbox{ if } x \leq \varepsilon, \\ 
x &\mbox{ if } x > \varepsilon, \\
\end{cases} 
\end{equation}
where $\varepsilon >0$. 
After applying modified ReLU, we know that sparsity increases from $15\%$ to about $35\%$ in the weights after pruning and then retraining, and in activation, applying modified ReLU (pruning) and then retraining will extend sparsity from $40\%$ to approximately $60-90\%$. It's clear that our modification will increase the sparsity to beyond $35\%$ and $90\%$ since the weights and activations that contribute negligible amounts still contribute negligible amounts after pruning; this is also because the sum of negligible numbers is still negligible relative to other weights and activations. This procedure will give us a swift identification of objects by machines as complexity has been further reduced by using modified ReLU. 

Therefore, we modify \eqref{eqn:modified-ReLU} and apply it to nonlinear functions like the sigmoid in \eqref{eqn:weighted-sum-activation-general} so that modified ReLU is applied to our deep neural network, i.e., 
\begin{equation}
\label{eqn:weighted-sum-activation-general-reduction}
\small 
\begin{aligned}
a_1^{(1)} 
&= 
\begin{dcases}
0 &\mbox{ if }  
    x_1^{(1)} \leq \varepsilon_1^{(1)}, \\ 
\sigma(x_1^{(1)} ) &\mbox{ if } 
    x_1^{(1)}  >  \varepsilon_1^{(1)}, \\
\end{dcases}   \\ 
a_2^{(1)} 
&=
\begin{cases}
0 &\mbox{ if }
x_2^{(1)}  \leq \varepsilon_2^{(1)}, 
\\ 
\sigma ( x_2^{(1)}) &
\mbox{ if } 
x_2^{(1)}  >  \varepsilon_2^{(1)}, \\ 
\end{cases} \\
\vdots &\qquad \qquad\qquad\qquad \vdots \\ 
a_{n_1}^{(1)} 
&= 
\begin{cases}
0 & \mbox{ if } 
x_{n_1}^{(1)} \leq \varepsilon_{n_1}^{(1)}, \\ 
\sigma(x_{n_1}^{(1)}) &\mbox{ if } 
x_{n_1}^{(1)}   >  \varepsilon_{n_1}^{(1)}, \\ 
\end{cases} \\ 
\vdots &\qquad \qquad\qquad\qquad \vdots \\ 
\end{aligned}
\end{equation}
where $\varepsilon_1^{(1)}, \varepsilon_2^{(1)},\ldots, \varepsilon_{n_1}^{(1)},\ldots \geq 0$, and 
\begin{align*}
    x_1^{(1)} &= w_{1,1}^{(0)}a_1^{(0)} + \ldots + 
w_{1,n_0}^{(0)}a_{n_0}^{(0)} + b_1^{(0)}, \\ 
    x_2^{(1)} &= w_{2,1}^{(0)}a_1^{(0)} +  
 \ldots + w_{2,n_0}^{(0)}a_{n_0}^{(0)} + b_2^{(0)}, \\  
 \vdots & \hspace{3cm}\vdots  \\ 
x_{n_1}^{(1)} &= w_{n_1,1}^{(0)}a_1^{(0)} + \ldots +  w_{n_1,n_0}^{(0)}a_{n_0}^{(0)} + b_{n_1}^{(0)}, \\ 
 \vdots & \hspace{3cm}\vdots  \\ 
\end{align*}

\section{Summary and Future Direction}
\label{section:summary-future-work}

Deep neural networks are emerging in many technologies today. 
We introduced three common neural networks and then discussed the role of activations, weights, hidden layers, and bias. Understanding the mathematics behind propagation and backpropagation enabled us to introduce an optimization model which further reduces the computational complexity by introducing more sparsity into the model. 
 With modified ReLU implemented in the deep neural network, we have enabled for the training and inference speeds to accelerate further on computing platforms consuming low computational resources.

\bibliography{model-reduction}

\end{document}